# Dynamic Model of Facial Expression Recognition based on Eigen-face Approach

Nikunj Bajaj
Department of Electrical Engineering
Indian Institute of Technology, Kharagpur
India – 721302

S L Happy, Aurobinda Routray
Department of Electrical Engineering
Indian Institute of Technology, Kharagpur
India – 721302

*Abstract*—Emotions are best way of communicating information; and sometimes it carry more information than words. Recently, there has been a huge interest in automatic recognition of human emotion because of its wide spread application in security, surveillance, marketing, advertisement, and human-computer interaction. To communicate with a computer in a natural way, it will be desirable to use more natural modes of human communication based on voice, gestures and facial expressions. In this paper, a holistic approach for facial expression recognition is proposed which captures the variation in facial features in temporal domain and classifies the sequence of images in different emotions. The proposed method uses Haar-like features to detect face in an image. The dimensionality of the eigenspace is reduced using Principal Component Analysis (PCA). By projecting the subsequent face images into principal eigen directions, the variation pattern of the obtained weight vector is modeled to classify it into different emotions. Owing to the variations of expressions for different people and its intensity, a person specific method for emotion recognition is followed. Using the gray scale images of the frontal face, the system is able to classify four basic emotions such as happiness, sadness, surprise, and anger.

I. INTRODUCTION

Emotion recognition systems have been grabbing high attention of the researcher and the industrialist communities since last two decades. It has been an interesting and challenging area, finding immense real life applications. Emotion recognition is widely being used to improvise the Human Computer Interaction. Speech and facial expressions, being most informative cues for emotion recognition, are extensively used in these systems. Facial expression analysis gained interest with recent advancements in face detection, face tracking, and face recognition techniques. This paper presents a dynamic model of emotions based on a holistic eigenspace based approach. Eigen space is a feature space that best encodes the variation in the eigenfaces. The eigenfaces may be considered as a set of features that characterize the global variation among face images.

A large number of emotion recognition algorithms have been proposed and implemented but most of them follow feature based method, where emotion recognition relies on localization and detection of facial features such as eyes, nose, mouth, chin, eyebrows and their geometrical relationships, with the help of deformable templates and extensive mathematics. B. Fasel and J. Luettin [1] described different types of approaches for facial expression analysis based on modeling of emotion, local feature extraction and motion extraction methods. However, classification of emotions through spatial information from static images is complicated. Xiang et al. [2] have extracted features using Fourier transform to represent an expression which is further processed to train a spatio-temporal model for each expression type using the fuzzy C means computation. In [3], Cohen *et al.* have used True Augmented Naive Bayes (TAN) classifiers to find dependencies among different facial expressions and Hidden Markov Model (HMM) to recognize emotion in temporal domain. Wang *et al.* [4] proposed real time facial expression recognition using AdaBoost technique in which they have first detected face using Haar features, then recognize six basic expressions from the features extracted from the detected face region. Applying PCA to local regions such as eyes and lips, D. Lin [5] has successfully classified the facial expressions using a Hierarchical Radial Basis Function Network (HRBFN). However, the accuracy of this method relies on the accurate detection of features like lips and eyes, which in itself is a difficult task. PCA has also been used in [6] to classify facial expressions from static images by extracting local features called Local Binary Pattern (LBP) [7]. In [8], sets of simple geometrical features derived from automatically detected and tracked facial feature point are used to segment a facial action into its temporal phases. Hybrid SVM and Hidden Markov Model (HMM) classifiers are used to model the time series. The proposed method in [9] uses facial velocity information over identified regions using optical flow and classifies the motion signatures to different emotions using Support Vector Machines (SVM) in real time. Attempt to track facial features using Gabor-wavelets and classifying them using Hierarchical Hidden Markov Models (HHMMs) has been reported in [10]. They have achieved 90% classification accuracy using Cohn-Kanade database. In [11], facial features are extracted using Active Appearance Model (AAM), which are further classified using SVM. However, shape-fitting models like Active Shape model (ASM) and AAM requires huge memory and have higher time complexity. However, for real time implementation of such algorithms, time complexity plays a crucial role, which is usually traded off with accuracy. Facial Action Coding System (FACS), developed by Ekman and Keltner, categorizes facial expressions in terms of Action Units (AUs). Another attempt to classify emotions in temporal domain is reported by [12] which uses multilevel HMM architecture to recognize emotions from videos in real time. However, detection of action units is a very difficult job, which decreases the overall recognition rate. In [13], authors have proposed a facial expression recognition algorithm where the emotion is classified according to the reconstruction error after the projection of each still image into orthogonal basis directions of different emotion subspaces.

In this paper, we present a holistic approach in which we identify different emotions based on the curve traced by successive facial features in temporal domain. The novelty of the proposed work is the use of projected weight vector into the eigenface subspace and observing its variation pattern with time for classification of the emotions into different categories. A standard face detection method, Haar classier, is used for detection of face in the image followed by PCA based dimensionality reduction. PCA yields better result in dimensionality reduction if the variance in data is minimal. Therefore, to achieve minimum variance in training and testing set, a dedicated database was created which comprises of a set of 50 images covering all expressions of interest. There is a sequence of 8 images for each emotion in one set. The database is divided into two categories: 40% for training and 60% for testing.

The rest of the paper is organized as follows. Section II describes the approach adapted for emotion recognition using an evolutionary model of human expression. In section III, we describe the results followed by the conclusion and future work in Section IV.

## II. METHODOLOGY

### A. Face detection

The foremost objective of the algorithm being the face detection, we have used Haar-classifier to detect face in an image owing to its high detection accuracy and real time performance [13]. Haar-like features are rectangular features in a digital image, which are used in object detection. Combination of such features in a cascade generates a strong classifier out of the weak classifiers. Fig. 1 shows some Haar-like features. The value of a rectangular Haar-like feature is the difference between the sum of the pixels within the white rectangular regions and that within the black rectangular regions.

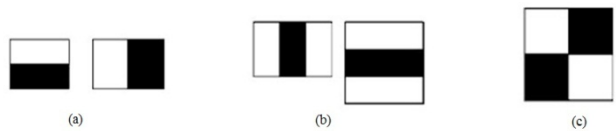

Fig. 1(a) Two-rectangle features (b) Three-rectangle features (c) Four-rectangle features

For fast computation of rectangular Haar-like features, integral images are computed (Fig. 2). Each element of the integral image contains the sum of all pixels located on the up-left region of the image. Using this, we compute the sum of rectangular areas in the image, at any position or scale. The integral image $I(x, y)$ is computed efficiently in a single pass over the image using:

$$I(x, y) = i(x, y) + I(x - 1, y) + I(x, y - 1) - I(x - 1, y - 1) \qquad (1)$$

In (1), $I(x, y) = \sum_{\substack{x' \leq x \\ y' \leq y}} i(x', y')$ and $i(x, y)$ is the intensity at point $(x, y)$.

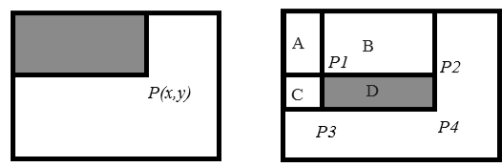

Fig. 2 Integral image representation

Once integral image is calculated, every feature can be obtained by using just 3 additions, irrespective of the scale and location. The number of Haar features available from a small image is very large. The optimal set of features and their corresponding thresholds for classification is obtained from AdaBoost algorithm.

For eigenface detection it is very important to normalize the image to line up the facial features and resample at the same pixel resolution. Therefore, scaling of the detected faces to a specified size was carried out after the face detection. Using the Haar-classifier for eyes, both the eyes were detected and aligned horizontally. By fixing the positions of the eyes in the detected face, face images were aligned.

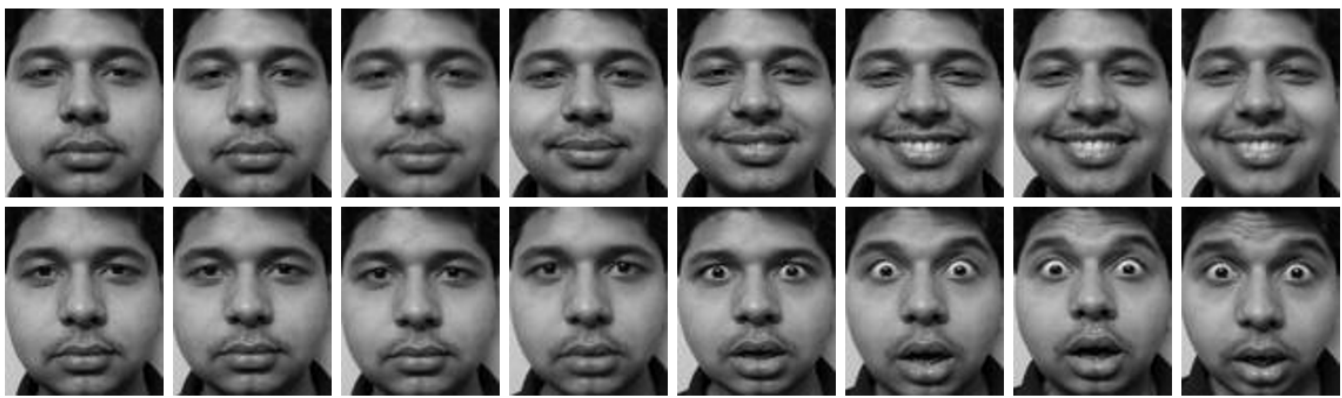

Fig. 3 Images used for constructing Eigenspace

## B. Dimensionality reduction using Principal Componect Analysis (PCA)

Eigenfaces are the principal components of a distribution of faces, or equivalently, the eigenvectors of the covariance matrix of the set of face images, where an image with N pixels is considered a point (or vector) in N-dimensional space. But images are high dimensional signal and dealing with eigenspace of the dimension of image vector can make the computation very expensive. Therefore, we use Principal Components Analysis (PCA) to reduce the dimension of data using the dependencies between the feature vectors. This helps to denote it in a more tractable form, without losing too much information. PCA seeks to find k principal axes, which define an orthonormal coordinate system that can capture most of the variance in data. Then each face image is approximated using a subset of the eigenfaces, those associated with the largest eigenvalues. Hence, the face is represented by the weighted sum of eigenfaces. The weights in different eigen directions are used in further classification process.

The eigenfaces are generated in the training session of PCA, in which the aligned training faces were used. A dedicated database was created for forming the eigenspace which, in our case, comprise of 50 images of faces covering all expressions of interest. All the eigenfaces were cropped to obtain the face and then scaled to the same size. Using covariance matrix of these 50 images an $MXN$ dimension space was constructed, where M and N are the number of rows and columns of the face image respectively. It can then be reduced to $K$ dimensions, using PCA. The number $K$ should be large enough for the construction of eigenspace without losing considerable information. A sample of 16 images from the dataset is Fig. 3. The procedure followed for eigenface retrieval is discussed in Table 1.

TABLE I.  TRAINING ALGORITHM

1. Each face images is transformed to vectors such as $\Gamma_1, \Gamma_2, ..., \Gamma_n$ of dimension $(h*w, 1)$ where $h$ and $w$ are the height and width of face image respectively
2. Mean feature vector is computed using
$$\psi = \frac{1}{M}\sum_{i=1}^{M}\Gamma_i$$
3. Mean feature vector is subtracted from each feature vector $\Gamma_i$
$$\phi_i = \Gamma_i - \psi_i$$
4. The covariance matrix $C$ is estimated using the equation
$$C = \frac{1}{M}\sum_{i=1}^{M}\phi_i\phi_i^T = AA^T$$
Where $A = [\phi_1, \phi_2..., \phi_M]$ of dimension $hw$ X $M$. Computing SVD of this large matrix is computationally difficult. Therefore, alternative method for finding eigen directions is implemented. $A^TA$ is computed instead of $AA^T$ as $M \ll hw$
5. The eigenvectors $v_i$ of $A^TA$ are computed by
$$\sigma_i u_i = Av_i$$
Hence the eigen-vectors $u_i$ of $AA^T$ are obtained.
6. Keep only $K$ eigenvectors corresponding to the $K$ largest Eigen values from each class (suppose $U = [u_1, u_2, ..., u_k]$)
7. All the $K$ eigenvectors are normalized.

## C. Projection of Test Image to EigenSpace

After obtaining the eigenfaces, the test image is expressed as a linear combination of these orthogonal images. It is similar to obtaining the components of the test image in the direction of different eigenvectors or simply plotting the image (which is just a point in the eigenspace) in the eigenspace. We thereby get a weight vector of K elements for each image by the following equation
$$w = \gamma^T * I$$
where, $w$ is the weight matrix, $\gamma$ is the matrix made of eigen face vectors and $I$ is the test image vector. The matrix $\gamma$ is of dimension $hwX50$ where $h$ and $w$ are the height and width of the eigenfaces respectively, 50 is the number of eigenfaces, and I is a vector of dimension $hw$ X 1. So this product gives us a vector of length $KX1$ that is the weight matrix for the test case. Since we have a sequence of 8 images characterizing each emotion we obtain a matrix of dimension 50X8 for each emotion. Therefore this stage generates a weight matrix for each of the desired emotions.

## D. Selection of Discriminating Directions

The previous stage generates weight matrix for different emotions, which in itself is of very high dimension and comparison of weight matrices of such dimension is difficult. Moreover, not all the eigen directions are relevant for discriminating the emotions. So, in the next step those directions are extracted which are most significant in identifying the emotion, hence the name discriminating direction. This was carried out by comparing the weight vectors along each eigen direction. The weight vectors for different emotion classes along the eigen directions having maximum variance were selected as discriminating directions. A proximity matrix, generated from the weight vectors, was used for choosing the weight vectors with maximum variances. Out of these, the 10 directions with maximum variance are chosen as the discriminating directions. The number 10 is actually obtained from a tradeoff between the performance and computation. The eigenfaces that correspond to 10 most significant directions is shown in Fig. 4.

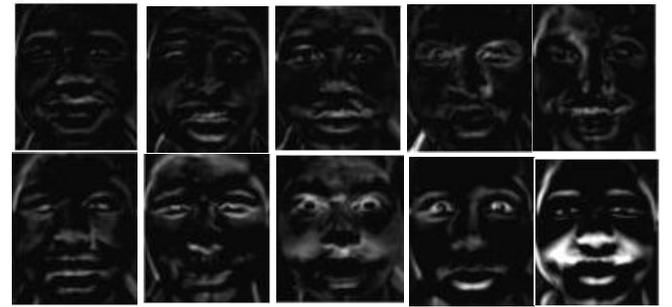

Fig. 4 Ten most significant Eigenfaces

## E. Curve Fitting

As discussed earlier, we have used a sequence of 8 images for each emotion in our dataset. After projection of the face image into the discriminating eigen directions, a weight vector corresponding to that face image is obtained. Similarly, projection of the sequence of images produces a sequence of weight vectors. The trajectory of the image sequence is captured for different emotions by taking the temporal distribution of the weight vectors of image sequence in the space of reduced dimension. The image sequence representing a particular emotion is characterized as a curve that is traced on the eigenspace. Therefore, a tentative trajectory of different emotions is obtained. A new set of image sequences is then classified to one of the emotion by selecting the emotion trajectory closer to the trajectory of the input image sequence.

TABLE II. POLYNOMIAL COEFFICIENTS

| P1 | P2 | P3 | P4 | P5 | P6 | P7 | P8 | P9 | P10 |
|---|---|---|---|---|---|---|---|---|---|
| 1 | 1 | 1 | 1 | 1 | 1 | 1 | 1 | 1 | 1 |
| 3.173 | -0.132 | -2.829 | -0.533 | 2.844 | -0.32 | -1.54 | 0.985 | -1.56 | -1.083 |
| 4.301 | -0.009 | 3.481 | -0.031 | 3.184 | -0.0113 | 0.952 | 0.325 | 0.973 | 0.425 |
| 3.256 | 0.002 | -2.434 | 0.057 | 1.768 | 0.062 | -0.299 | 0.037 | -0.300 | -0.072 |
| 1.507 | -6.8E-05 | 1.057 | -0.010 | 0.498 | -0.009 | 0.05 | -0.001 | 0.043 | 0.004 |
| 0.4373 | 1.7E-07 | -0.292 | 0.0006 | 0.062 | 0.0003 | -0.004 | -0.0004 | -0.0007 | 0.0002 |
| 0.0777 | -6.8E-08 | 0.0501 | 3.2E-07 | 0.0018 | 3.0E-05 | 0.0001 | -8.7E-06 | -0.0005 | -2.6E-05 |
| 0.0077. | -6.5E-11 | -0.0049 | -7.8E-07 | -0.0001 | -1.8E-06 | 3.0E-07 | 9.4E-07 | 3.6E-05 | 5.2E-07 |
| 0.0003 | -1.3E-12 | 0.0002 | 1.5E-08 | 7.8E-07 | 1.3E-08 | 1.75E-10 | -4.59E-09 | 2.8E-07 | 1.1E-08 |

Considering the fact that we are dealing with high dimensional data and we have just 8 sample points for each sequence, an implicit equation is difficult to obtain. If we make a polynomial approximation then even a 3rd order approximation requires 220 coefficients, which is impossible to obtain given the information. So we fit separate curves along each discriminating direction. An 8th order polynomial can be obtained along each direction since we know 8 roots. Therefore a table is obtained for each emotion of interest which represents the coefficients of polynomials along all 10 directions as shown in Table II.

For instance, the first polynomial will be of the form –

$$x^8 + 3.17x^7 + 4.3x^6 + 3.25x^5 + 1.5x^4 + 0.43x^3 + 0.077x^2 + 0.007x + 0.0003 = 0$$

Here, the dimension of this matrix is $9X10$. Since we are operating on 10 major directions and in each direction we have a polynomial of 8th order, which is defined by 9 coefficients. Once we obtain these equations, we obtain the error of all the data points in the input sequence with respect to the polynomials. If data points are checked with corresponding polynomials then error obtained is very less.

III. RESULTS

The above algorithm has been tested and verified to work well with the emotions of - Happiness, Sorrow, Surprise and Anger. As mentioned earlier, we obtain polynomial equations for each of the 10 major directions for each sequence of images representing a particular expression. The best fit polynomial is further obtained by using many such training data sets. For all these data sets a polynomial is obtained out of which the best fit is obtained by using Least Square Error. The test sequence data points are supposed to satisfy these equations if the emotions match else they should deviate from zero. Therefore, error should be least if the weight matrix of a particular emotion is compared with the polynomial corresponding to same expression. With this idea we generate the confusion matrix for all the desired emotions. Out of a total of 40 datasets 24 were used for testing and the matrix obtained is shown in Table III. The numbers in the matrix represent the fraction of number of sequence that match with the polynomial corresponding to the particular emotion with respect to the total number of test cases.

TABLE III. CONFUSION MATRIX

|  | Anger | Happiness | Sorrow | Surprise |
|---|---|---|---|---|
| Anger | 0.833 | 0 | 0.125 | 0.041 |
| Happiness | 0.041 | 0.916 | 0 | 0.041 |
| Sorrow | 0.208 | 0 | 0.791 | 0 |
| Surprise | 0.041 | 0.125 | 0 | 0.833 |

We see that the diagonal elements are much greater in magnitude than the off-diagonal elements representing that a large fraction of emotions are correctly identified. On an average the algorithm works with an accuracy of 84.4 %.

IV. CONCLUSION

This paper briefly overviews automatic emotion recognition using a novel holistic approach in which we identify different emotions based on the curve traced by extracted features of the sequence of face images in the eigen-space. Limitations of spatial domain pattern recognition is complemented by features used in temporal domain and satisfactory results are obtained in emotion recognition of the person when training is carried out by images of the same person, i.e. by using the customized eigen subspace created by using the images of that person. Interesting results can be obtained by using similar technique in emotion recognition for different people after being trained by a generic dataset. Moreover, the results can be further improved by introducing machine learning for training the polynomial coefficients. Research in the domain of emotion recognition has attained some success, but is still an area to be explored further.

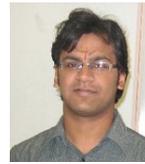

**Nikunj Bajaj** has received B.Tech. Hons. in Instrumentation Engineering from Indian Institute of technology, Kharagpur in 2013. He is currently pursing MS in the Department of Electrical Engineering and Computer Sciences in University of California, Berkeley. His research interests are signal processing, s design and optimization of safety critical cyber-physical systems, modeling and Analysis of interoperability of heterogeneous systems.

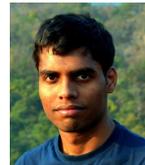

**S L Happy** has received the B.Tech. (Hons.) degree from Institute of Technical Education and Research (ITER), India in 2011. Now he is pursuing the M. S. degree from Indian Institute of Technology (IIT) Kharagpur, India. His research interests include pattern recognition, computer vision and facial expression analysis.

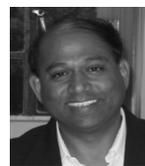

**Aurobinda Routray** is a professor in the Department of Electrical Engineering, Indian Institute of Technology, Kharagpur, India. His research interest includes non-linear and statistical signal processing, signal based fault detection and diagnosis, real time and embedded signal processing, numerical linear algebra, and data driven diagnostics.